\documentclass[pdflatex,sn-mathphys-num]{sn-jnl}% Math and Physical Sciences Numbered Reference Style
\usepackage{graphicx}%
\usepackage{multirow}%
\usepackage{amsmath,amssymb,amsfonts}%
\usepackage{amsthm}%
\usepackage{mathrsfs}%
\usepackage[title]{appendix}%
\usepackage[table,xcdraw]{xcolor}
\usepackage{textcomp}%
\usepackage{manyfoot}%
\usepackage{booktabs}%
\usepackage{algorithm}%
\usepackage{algorithmicx}%
\usepackage{algpseudocode}%
\usepackage{listings}%
\usepackage{booktabs}
\usepackage{multirow}
\usepackage{graphicx}
\usepackage[normalem]{ulem}
\usepackage{arydshln}
\usepackage{bm}
\usepackage{pifont}
\usepackage{enumitem}
\useunder{\uline}{\ul}{}
\definecolor{Blue}{RGB}{175, 215, 255}
\theoremstyle{thmstyleone}%
\theoremstyle{thmstyletwo}%

\theoremstyle{thmstylethree}%

\begin{document}

\title[MaterEval]{From Blind Guess to Informed Judgment: Teaching LLMs to Evaluate Materials by Building Knowledge-Augmented Preference Signals}


\author[1]{\fnm{Yeyong} \sur{Yu}}\email{yuyeyong@shu.edu.cn}

\author[1]{\fnm{Wenya} \sur{Hu}}\email{wenyahu@shu.edu.cn}

\author[1]{\fnm{Xing} \sur{Wu}}\email{xingwu@shu.edu.cn}

\author*[1,2,3,4]{\fnm{Quan} \sur{Qian}}\email{qqian@shu.edu.cn}

\affil[1]{\orgdiv{School of Computer Engineering \& Science}, \orgname{Shanghai University}, \orgaddress{ \city{Shanghai}, \postcode{200444}, \country{China}}}

\affil[2]{\orgdiv{Center of Materials Informatics and Data Science, Materials Genome Institute},
	\orgname{Shanghai University}, \orgaddress{ \city{Shanghai}, \postcode{200444}, \country{China}}}

\affil[3]{\orgdiv{Key Laboratory of Silicate Cultural Relics Conservation (Shanghai University)}, \orgname{Ministry of Education}, \country{China}}

\affil[4]{\orgdiv{Shanghai Institute for Advanced Communication and Data Science},
	\orgname{Shanghai University}, \orgaddress{ \city{Shanghai}, \postcode{200444}, \country{China}}}


\abstract{As candidate generation and high-throughput experimentation advance, the primary bottleneck in materials discovery is shifting from \emph{property prediction} to making \emph{reliable evaluations} among massive candidate sets. 
We propose a Knowledge-Augmented Preference Signals Framework \textsc{MaterEval} that automatically produces, for the same candidate, two evaluations: an \emph{informed judgment} that follows expert rules and provides supporting evidence, and a rule-removed \emph{blind guess}. 
By pairing the two evaluations as preference data, we guide general-purpose Large Language Models (LLMs)—originally lacking materials-specific criteria—from intuitive judgment toward \emph{reliable evaluation} supported by explicit evidence.
To balance throughput, cost, and reliability, we further introduce a fast--slow reasoning scheme that decouples large-scale rapid screening from in-depth review on a small subset.
Using high-entropy alloy (HEA) assessment as a case study, we show that, without external retrieval and relying solely on internalized capabilities, small open-source LLMs achieve substantial gains in accuracy, conclusion consistency, and evidence discrimination, approaching the performance of rule-based closed-source LLMs. These results demonstrate that expert rules can be systematically transformed into learnable preference signals, enabling a low-cost and deployable evaluation module for autonomous materials discovery loops.}




\maketitle

\section*{Main}
\label{sec:intro}
\setcounter{section}{1}

In recent years, mainstream progress in AI for Science within the materials domain has largely focused on \emph{materials property prediction}: given a composition or microstructure, machine learning models or large language models are used to predict key properties \citep{merchant2023scaling, yu2023better, liang2025review}, thereby accelerating the initial screening of candidate materials. 
While such approaches have achieved notable success in learning mappings from data to properties, they remain fundamentally driven by statistical correlations and in-distribution interpolation. 
As a result, they often lack explicit incorporation of domain knowledge and expert experience required for assessing \emph{feasibility} and \emph{reliability} \citep{madika2025artificial, jiang2025interpretable, yu2024small}. 

Consequently, accurate property prediction does not necessarily translate into reliable guidance on material synthesizability. 
Even when predicted values appear promising, they may merely reflect statistical fits to noisy or biased data distributions, without sufficient consideration of processing constraints, mechanistic consistency, or supporting evidence. 
Such limitations can expose substantial risks in critical aspects such as phase stability, synthesis pathways, and reproducibility \citep{jacobs2024machine, bartel2020critical}. 
In other words, current AI systems are effective at answering the question, \emph{``what properties might this composition have?''}, but remain far less capable of addressing \emph{``can we actually make it, and should we trust this data or design?''}.

With the emergence of AI laboratories and autonomous discovery platforms \citep{tobias2025autonomous, zaki2025self, hung2024autonomous}, this tension is further amplified. 
As candidate materials can be generated by models at extremely high throughput and experiments executed in parallel by automated systems, the limiting factor of closed-loop efficiency is gradually shifting from \emph{prediction speed} to the \emph{reliability of evaluation} \citep{amirian2025building}. 
This includes assessing the credibility of materials data itself (\emph{data quality assessment}), as well as comprehensively evaluating the feasibility and risks of synthesis plans (\emph{synthesis feasibility} or \emph{preparation potential}) \citep{reeves2026data, nematov2025machine}. 

Such judgments rely on dense, coupled knowledge across composition, processing, and properties, and require careful verification of supporting evidence chains. 
In practice, they are typically performed by domain experts, making them both knowledge-intensive and costly, and therefore difficult to scale linearly with the growing number of candidates \citep{tom2024self, jiang2025ai4materials}. 
Consequently, as materials research moves from property prediction toward autonomous discovery, a critical gap becomes apparent: there is a need for a scalable and low-cost model capability that can fully leverage domain knowledge to filter a vast pool of \emph{apparently promising} predictions into a small set of candidates that are truly \emph{worth pursuing}, \emph{feasible to realize}, and \emph{reliable in outcome}—shifting materials discovery from blind guess to informed judgment.

\begin{figure}
	\centering
	\includegraphics[width=1\textwidth]{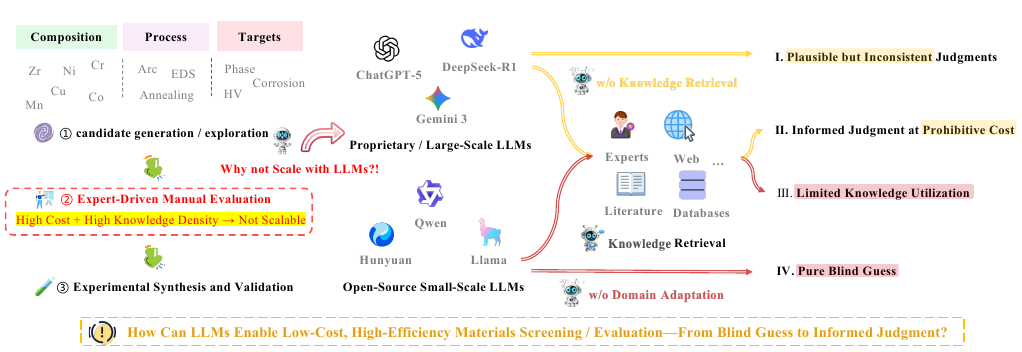}
	\caption{\textbf{Motivation and limitations of LLM-based materials evaluation.} 
\textbf{Left:} The canonical materials discovery workflow commonly adopted in practice. A large number of candidate materials are first generated in a vast design space (\ding{172}), followed by expert-driven manual evaluation (\ding{173}), after which only a small subset of candidates is selected for experimental synthesis and validation (\ding{174}). While candidate generation can be readily scaled, expert evaluation remains a high-cost, high–knowledge-density bottleneck that limits the overall throughput of the discovery loop. 
\textbf{Right:} Representative paradigms for applying LLMs to materials evaluation and their inherent limitations, including 
(I) plausible but inconsistent judgments without knowledge support, 
(II) prohibitive cost when relying on large closed-source models with retrieval, 
(III) limited knowledge utilization in small models augmented with external knowledge, and 
(IV) degeneration into blind guess under strict efficiency constraints.}
\label{fig:intro}
\end{figure}

To make this limitation explicit, Fig.~\ref{fig:intro} (left) summarizes the canonical materials discovery pipeline that remains prevalent in practice. 
A large number of candidates are first generated in a vast materials design space (\ding{172}~candidate generation / exploration), followed by a critical decision stage dominated by \emph{expert-driven manual evaluation} (\ding{173}), after which only a small subset of candidates is selected for \emph{experimental synthesis and validation} (\ding{174}). 

As illustrated, the central bottleneck of this workflow does not arise from insufficient candidate generation, but from the intrinsic characteristics of stage~\ding{173}. Effective evaluation requires the integration of literature knowledge, processing constraints, and mechanistic consistency, and relies heavily on tacit expertise that is difficult to standardize or parallelize. 
As the scale of generated candidates continues to grow, this expert-driven step increasingly limits the throughput of the discovery loop, preventing many apparently promising predictions from being translated into synthesizable, reliable, and actionable experimental decisions.

In recent years, large language models (LLMs) have demonstrated strong potential in scientific reasoning and knowledge integration \citep{petrosanu2023tracing, rane2023contribution}, opening new possibilities for automating materials evaluation \citep{mishra2024foundational, bajan2025exploring}. 
Against this backdrop, a natural question arises for stage~\ding{173}: \emph{why not scale expert evaluation with LLMs?} 
Specifically, one would like models to assess the credibility of materials data (\emph{data quality assessment}) and to evaluate the feasibility and potential risks of candidate synthesis plans (\emph{synthesis feasibility} or \emph{preparation potential}), thereby enabling scalable screening and prioritization as the candidate space rapidly expands.

However, applying LLMs to materials evaluation remains fundamentally challenging. As summarized in the right panel of Fig.~\ref{fig:intro}, existing efforts can be broadly categorized into four representative paradigms, each of which suffers from critical limitations that are difficult to circumvent:

\begin{enumerate}[label=(\roman*)]
\item First, directly applying general-purpose LLMs to materials evaluation without external knowledge retrieval often leads to \emph{plausible but inconsistent} outcomes. Although the models can generate superficially reasonable explanations, they lack explicit grounding in materials mechanisms, processing constraints, and experimental context. As a result, evaluations may vary substantially across different samples, or even across repeated assessments of the same sample, making it difficult to establish stable and reliable evaluation criteria.

\item Second, combining large closed-source LLMs with knowledge bases or retrieval-augmented mechanisms \citep{lewis2020retrieval, xi2025survey} can substantially improve the quality of single evaluations. However, in realistic materials screening workflows, this paradigm typically suffers from prohibitive cost. Frequent retrieval, long-context reasoning, and repeated calls to closed-source models cause evaluation costs to scale rapidly with the number of candidates, rendering this approach impractical for large-scale exploration and closed-loop experimentation.

\item Third, to reduce cost, one may resort to small open-source LLMs augmented with external knowledge bases. In practice, however, such models are often constrained by limited capacity, resulting in \emph{limited knowledge utilization}. Although relevant information can be retrieved, the models struggle to effectively integrate heterogeneous sources, weigh conflicting evidence, and produce consistent evaluations.

\item Finally, under strict efficiency and cost constraints that preclude sufficient domain adaptation, small general-purpose LLMs tend to degenerate into \emph{pure blind guess} when applied to materials problems. In this regime, models neither perform effective screening nor provide reliable evaluation signals, a limitation that is particularly pronounced in high--knowledge-density tasks such as materials data quality assessment and synthesis feasibility analysis.
\end{enumerate}

Taken together, these limitations indicate that directly applying LLMs to materials evaluation does not naturally yield a transition from \emph{blind guess} to \emph{informed judgment}. 
The fundamental bottleneck lies not in insufficient model scale, but in the absence of a training mechanism that can systematically inject domain knowledge into LLMs and align their judgments in a scalable and reliable manner.

To address these challenges, we propose a knowledge-augmented preference alignment framework, \textsc{MaterEval}, designed to train small, open-source LLMs (e.g., Qwen \citep{qwen} and Hunyuan \citep{hunyuan}) to perform materials evaluation at a level comparable to large closed-source models (e.g., GPT-5 and DeepSeek-R1 \citep{deepseek}). 
The evaluation targets considered in this work include, but are not limited to, assessing the credibility of materials data and analyzing the synthesis feasibility of candidate material designs. 
Although these tasks differ at the application level, they share a common formulation: given a set of composition, processing conditions, and target properties, the model is required to perform a holistic evaluation of the plausibility and feasibility of a material candidate.

To bridge the capability gap of small-parameter models in such high--knowledge-density tasks, we introduce the core idea of \emph{knowledge-augmented preference signals}. 
Rather than training models merely to generate plausible explanations, our objective is to enable them to form \emph{stable, repeatable, and aligned judgments} in materials-specific contexts. 
Concretely, we abstract dispersed expert knowledge in materials evaluation into actionable knowledge bases, and further construct preference signals that distinguish candidates that are \emph{more credible or synthesizable} from those that are \emph{superficially plausible yet fundamentally unreliable}. 

By learning from these preference signals, small open-source LLMs can acquire materials evaluation capabilities approaching those of powerful closed-source models, while operating under controlled and scalable cost. 
This process establishes a closed training loop that transitions from rule-driven knowledge injection to preference-level judgment alignment, allowing small LLMs to gradually develop stable and generalizable materials evaluation capabilities without relying on expensive expert annotations.

Our main contributions are summarized as follows:

\begin{enumerate}
\item \textbf{Knowledge-augmented preference signals for materials evaluation.} 
We propose a fully automated and low-cost pipeline to transform dispersed domain knowledge in materials evaluation into learnable preference signals, enabling models to distinguish reliable, evidence-supported judgments from superficially plausible but unreliable ones.

\item \textbf{Stage-wise judgment alignment for small open-source LLMs.} 
We introduce a two-stage training framework that aligns small open-source LLMs with expert evaluation criteria through rule-derived supervision and preference-based judgment alignment, substantially improving the stability and reliability of their evaluation outputs.

\item \textbf{Hierarchical fast--slow evaluation balancing efficiency and reliability.} 
We present a hierarchical evaluation paradigm that decouples high-throughput screening from in-depth assessment, achieving scalable materials evaluation with reduced inference cost while preserving consistency and credibility.
\end{enumerate}

Finally, we validate the effectiveness of the proposed framework through systematic evaluation. 
Under a rule-free setting (without explicit knowledge support), the trained small-parameter models achieve stability and consistency comparable to rule-based evaluations produced by powerful LLMs such as GPT-5 and DeepSeek, while exhibiting well-aligned evaluation outcomes. 
These results demonstrate that, by introducing knowledge-augmented preference signals together with an appropriate alignment mechanism, small models can reliably assume key judgment roles in materials evaluation and screening. 
This capability enables a practical and scalable decision-making module for AI-lab-driven materials discovery workflows.

\section*{Results}
\label{sec:results}
\setcounter{section}{2}
\setcounter{subsection}{0}

\subsection{Overview of Knowledge-Augmented Preference Alignment for Materials Evaluation
}
\label{framework}

\begin{figure}
	\centering
	\includegraphics[width=1\textwidth]{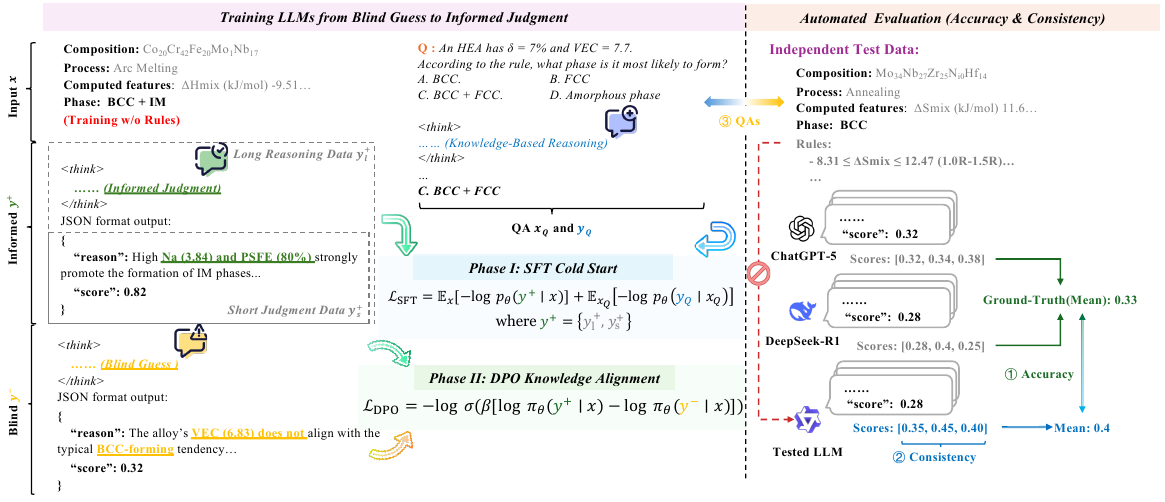}
	\caption{\textbf{Knowledge-augmented preference alignment framework} \textsc{MaterEval} \textbf{for materiasl evaluation.}
\textbf{Left:} Two-stage alignment of LLMs in a rule-free setting, where the input $x$ consists only of composition, processing conditions, and computed material descriptors. Phase~I performs an supervised fine-tuning (SFT) cold start using (i) long-form reasoning outputs $y_l^{+}$, (ii) concise judgment outputs $y_s^{+}$, and (iii) auxiliary rule-derived QAs $y_Q$. Phase~II applies direct preference optimization (DPO) to favor knowledge-grounded judgments $y^{+}$ over rule-removed blind guesses $y^{-}$. 
\textbf{Right:} Automated evaluation on an independent test set jointly measures accuracy (alignment between the predicted mean score and the ground-truth mean), consistency (dispersion across repeated stochastic generations), and QA accuracy (correctness on rule-derived questions probing domain knowledge internalization).
}
\label{fig:framework}
\end{figure}

To address the limited scalability of materials evaluation in large candidate spaces, we propose a Knowledge-Augmented Preference Alignment Framework \textsc{MaterEval}  (Fig.~\ref{fig:framework}) that injects domain expert knowledge into small open-source LLMs in a low-cost and deployable manner. 
The core idea of the framework is to automatically construct two types of evaluation signals for the same candidate material: an \emph{informed judgment} that follows expert rules and is supported by explicit evidence, and a \emph{blind guess} that is generated without rule guidance and relies solely on linguistic priors. 

By pairing these two signals as preference data for training, the model is systematically guided away from unstable intuitive judgments toward consistent, evidence-grounded evaluation (the data construction process is detailed in \textsection~\ref{sec:dataset}). 
Importantly, domain knowledge is not introduced at inference time via external rules or retrieval prompts. 
Instead, it is internalized into the model parameters through supervised and preference-based training, enabling the model to perform materials evaluation without access to additional knowledge sources during inference. 
This design substantially reduces deployment cost while preserving evaluation reliability.

The training process adopts a two-stage alignment strategy. 
In the first stage, the model is cold-started via supervised fine-tuning (SFT). 
Teacher models provide demonstration samples that include explicit reasoning traces and structured conclusions, enabling the student model to learn a stable evaluation output format that integrates \emph{conclusion, supporting evidence, and concise summary}. 
In parallel, we introduce domain-specific knowledge question--answering (QA) tasks (binary and multiple-choice questions) automatically derived from rule bases as auxiliary supervision. 
These tasks reinforce the model’s alignment with key evaluation criteria, conceptual boundaries, and mechanistic principles, thereby improving the efficiency of domain knowledge injection. 

In the second stage, we apply direct preference optimization (DPO) to perform paired alignment between \emph{knowledge-grounded judgments} and corresponding \emph{blind guesses} under identical inputs. 
This process guides the model to consistently prefer the former in parameter space, further enhancing the robustness and consistency of evaluation outcomes (training details are provided in \textsection~\ref{sec:training}).

Building on this foundation, we introduce a fast--slow hierarchical evaluation paradigm to jointly address screening efficiency and evaluation reliability. 
This paradigm is explicitly modeled during training: each evaluation instance is decomposed into a short \emph{fast} judgment output for preliminary screening and a long-form \emph{slow} reasoning output for in-depth assessment. 
These outputs are differentiated using dedicated control templates, allowing the model to learn to produce judgments at different granularities under varying computational budgets. 
At deployment time, the model first operates in fast mode to perform high-throughput screening over large candidate sets, and then switches to slow mode for a small subset of high-potential candidates to generate more comprehensive, evidence-integrated evaluations. 
This design preserves reliability and consistency while maintaining controlled inference cost.

The right panel of Fig.~\ref{fig:framework} illustrates our automated evaluation protocol on an independent test set, designed to assess whether the model has acquired practically usable materials evaluation capabilities. 
We evaluate performance along three complementary dimensions (evaluation details are provided in \textsection~\ref{sec:evaluation}):
\begin{itemize}
	\item \emph{Accuracy} measures the alignment between the model’s predicted mean scores and aggregated reference judgments produced by strong teacher models (e.g., GPT-5 and DeepSeek-R1). 
	\item \emph{Consistency} characterizes the stability of model outputs across repeated independent evaluations of the same sample, which is a critical prerequisite for reliable ranking and decision-making.
	\item  \emph{QA Accuracy} assesses the model’s explicit mastery of evaluation rules and mechanistic knowledge through objectively verifiable questions, enabling us to distinguish genuine knowledge internalization from superficial score fitting.
\end{itemize}

Taken together, the proposed Knowledge-Augmented Preference Alignment Framework \textsc{MaterEval} enables low-cost transfer of materials evaluation knowledge to small open-source LLMs without relying on external retrieval or explicit rule prompting. 
Its reliability and deployability are systematically validated through converging evidence from accuracy, consistency, and knowledge mastery. 
The subsequent Results section quantitatively reports the performance gains achieved by the framework and analyzes the contribution of individual components through comparative and ablation studies.

\subsection{Experimental Results}
\label{sec:exp_results}

To systematically evaluate the reliability and stability of \textsc{MaterEval} on materials data quality assessment tasks, we design a comprehensive experimental protocol tailored for LLMs. 
Our evaluation focuses on two core dimensions: \emph{Accuracy} and \emph{Consistency}.
Accuracy measures the deviation between model predictions and a reference ground truth, reflecting how closely the model aligns with expert-level evaluation trends. 
Consistency characterizes the stability of model outputs across repeated independent evaluations of the same sample, capturing reproducibility and internal robustness. 
Jointly assessing these two dimensions provides a more complete characterization of the credibility of LLM-based materials data quality evaluation, and serves as a quantitative foundation for subsequent comparative and ablation analyses.

\vspace{0.2cm}
\noindent \textbf{Dataset and Evaluation Setup}
We select the high-entropy alloy (HEA) system as the evaluation scenario and construct test tasks around six representative materials properties: phase constitution (Phase), elongation (Elongation), ultimate tensile strength (UTS), microhardness (HV), corrosion resistance (Corrosion), and oxidation resistance (Oxidation). 
These properties span multiple levels from microstructural characteristics to macroscopic mechanical performance and environmental stability, enabling evaluation of model judgment capabilities across diverse physical mechanisms and assessment perspectives.

For each property evaluation task, we randomly sample 50 test instances from a standard evaluation set constructed based on automatically distilled assessment rules from the materials knowledge base. 
Under a unified evaluation protocol and identical input settings, GPT-5 and DeepSeek-R1 \citep{deepseek} independently assess each test instance three times. The six resulting scores from the two teacher models are averaged to form the reference ground truth for the corresponding property.

All other models are evaluated on the same test instances, with three independent evaluations conducted per sample. 
The deviation between each model’s predicted scores and the reference ground truth is used to compute the \emph{Accuracy} metric, quantifying alignment with expert-level evaluation trends.
Meanwhile, the dispersion among the three independent evaluations is used to compute the \emph{Consistency} metric, reflecting the stability of model judgments under identical input conditions.

We consider two categories of baseline models for comparison:
\begin{itemize}
\item \textbf{General baselines:} GPT-5, GPT-4o \citep{gpt4o}, DeepSeek-R1 \citep{deepseek}, Claude-Sonnet-4.5, and Gemini-2.5-Pro \citep{gemini}.
\item \textbf{Custom-trained models:} the Hunyuan-Instruct \citep{hunyuan} and Qwen3-Instruct \citep{qwen} series. The Hunyuan family includes models with 1.8B, 4B, and 7B parameters, while the Qwen3 series includes 1.7B, 4B, and 8B variants.
\end{itemize}

All models are evaluated using the same test samples and statistical procedures to ensure fair comparison and statistical validity across different architectures and parameter scales.

\vspace{0.2cm}
\noindent\textbf{Metrics}
To comprehensively evaluate the performance of LLMs on materials data quality assessment tasks, we consider two complementary dimensions: \emph{Accuracy} and \emph{Consistency}.

Accuracy measures how closely a model’s evaluation results align with the reference ground truth. We report four standard metrics:
\begin{itemize}
\item \textbf{MAE} and \textbf{RMSE} quantify the mean absolute error and root mean squared error, respectively, with lower values indicating more accurate predictions;
\item \textbf{Absolute Bias (Abs Bias)} measures systematic deviation, capturing whether a model’s evaluations are consistently overestimated or underestimated;
\item \textbf{$R^2$} (coefficient of determination) reflects the linear correlation between predicted scores and ground truth values, where values closer to 1 indicate better alignment.
\end{itemize}

Consistency characterizes the stability and reproducibility of model outputs across multiple independent evaluations of the same sample. We report four complementary metrics:
\begin{itemize}
\item \textbf{Std.} (standard deviation) measures the dispersion of scores across repeated evaluations;
\item \textbf{Agreement@5 (\%)} defines a tolerance threshold $\varepsilon = 5\%$. For repeated scores $\{s_j\}$ on the same sample, the evaluations are considered consistent if $\max(s_j) - \min(s_j) \le \varepsilon$, and inconsistent otherwise;
\item \textbf{Pairwise MAD} (mean absolute deviation) computes the average of $|s_i - s_j|$ over all score pairs for the same sample, providing a robust measure of dispersion that is less sensitive to outliers than variance or standard deviation;
\item \textbf{Krippendorff’s $\alpha$} quantifies overall agreement among repeated evaluations, with values closer to 1 indicating higher stability and stronger inter-evaluation reliability.
\end{itemize}

\begin{table}[htbp]
\centering
\renewcommand\arraystretch{1.5} 
\resizebox{\textwidth}{!}{
\begin{tabular}{@{}lccccccccc@{}}
\toprule
\multicolumn{1}{c}{\multirow{2}{*}{\textbf{Models}}} & \multirow{2}{*}{\textbf{Size}} & \multicolumn{4}{c}{\textbf{Accuracy}} & \multicolumn{4}{c}{\textbf{Consistency}} \\ \cmidrule(l){3-10} 
\multicolumn{1}{c}{} &  & \multicolumn{1}{c}{\textbf{MAE $\downarrow$}} & \multicolumn{1}{c}{\textbf{RMSE $\downarrow$}} & \multicolumn{1}{c}{\textbf{Abs Bias $\downarrow$}} & \multicolumn{1}{c}{\textbf{$\bm{R^2}$ $\uparrow$}} & \multicolumn{1}{c}{\textbf{Std. $\downarrow$}} & \multicolumn{1}{c}{\textbf{Agree@5(\%) $\uparrow$}} & \multicolumn{1}{c}{\textbf{Pair. MAD $\downarrow$}} & \multicolumn{1}{c}{\textbf{Kripp. $\bm{\alpha}$ $\uparrow$}} \\ \midrule
\multicolumn{10}{l}{{\ul \textit{\textbf{General Baselines}}}} \\

\textbf{gpt-5 \textit{(with know)}} & - &
\multicolumn{4}{c}{\multirow{2}{*}{N/A}} &
\cellcolor{Blue!100}\textbf{4.17} &
\cellcolor{Blue!100}\textbf{47.33} &
\cellcolor{Blue!100}\textbf{5.26} &
\cellcolor{Blue!100}\textbf{0.96} \\

\textbf{DeepSeek-R1 \textit{(with know)}} & 671B &
\multicolumn{4}{c}{} &
\cellcolor{Blue!50}5.00 &
\cellcolor{Blue!60}29.67 &
\cellcolor{Blue!50}6.27 &
\cellcolor{Blue!80}0.93 \\

\textbf{gpt-5} & - &
\cellcolor{Blue!45}11.27 ± 0.21 &
\cellcolor{Blue!45}15.42 ± 0.38 &
\cellcolor{Blue!45}11.27 ± 0.21 &
\cellcolor{Blue!50}0.60 ± 0.02 &
\cellcolor{Blue!80}4.70 &
\cellcolor{Blue!90}{\ul 42.67} &
\cellcolor{Blue!80}5.91 &
\cellcolor{Blue!60}0.92 \\

\textbf{Claude-sonnet-4.5} & - &
\cellcolor{Blue!40}13.39 ± 0.16 &
\cellcolor{Blue!40}16.65 ± 0.36 &
\cellcolor{Blue!40}13.39 ± 0.16 &
\cellcolor{Blue!40}0.53 ± 0.02 &
\cellcolor{Blue!35}6.43 &
\cellcolor{Blue!40}16.00 &
\cellcolor{Blue!35}8.15 &
\cellcolor{Blue!40}0.90 \\

\textbf{Gemini-2.5-pro} & - &
\cellcolor{Blue!25}15.16 ± 0.14 &
\cellcolor{Blue!25}18.93 ± 0.31 &
\cellcolor{Blue!25}15.16 ± 0.14 &
\cellcolor{Blue!30}0.39 ± 0.02 &
\cellcolor{Blue!15}8.57 &
\cellcolor{Blue!20}7.67 &
\cellcolor{Blue!15}10.91 &
\cellcolor{Blue!15}0.84 \\

\textbf{gpt-4o} & - &
\cellcolor{Blue!35}14.25 ± 0.32 &
\cellcolor{Blue!35}17.86 ± 0.5 &
\cellcolor{Blue!35}14.25 ± 0.32 &
\cellcolor{Blue!35}0.46 ± 0.03 &
\cellcolor{Blue!30}6.69 &
\cellcolor{Blue!30}13.33 &
\cellcolor{Blue!30}8.51 &
\cellcolor{Blue!35}0.89 \\

\textbf{DeepSeek-R1} & 671B &
\cellcolor{Blue!30}14.68 ± 0.09 &
\cellcolor{Blue!30}19.47 ± 0.2 &
\cellcolor{Blue!30}14.68 ± 0.09 &
\cellcolor{Blue!25}0.36 ± 0.01 &
\cellcolor{Blue!60}4.92 &
\cellcolor{Blue!50}28.00 &
\cellcolor{Blue!60}6.16 &
\cellcolor{Blue!20}0.85 \\

\textbf{Hunyuan-7B-Instruct} & 7B &
\cellcolor{Blue!15}18.77 ± 0.63 &
\cellcolor{Blue!15}23.2 ± 0.86 &
\cellcolor{Blue!15}18.77 ± 0.63 &
\cellcolor{Blue!15}0.09 ± 0.07 &
\cellcolor{Blue!10}17.16 &
\cellcolor{Blue!10}2.00 &
\cellcolor{Blue!10}21.8 &
\cellcolor{Blue!10}0.59 \\

\textbf{Qwen3-8B-Instruct} & 8B &
\cellcolor{Blue!20}18.26 ± 0.64 &
\cellcolor{Blue!20}22.58 ± 1.04 &
\cellcolor{Blue!20}18.26 ± 0.64 &
\cellcolor{Blue!20}0.13 ± 0.08 &
\cellcolor{Blue!5}17.65 &
\cellcolor{Blue!5}1.33 &
\cellcolor{Blue!5}22.39 &
\cellcolor{Blue!5}0.59 \\

\midrule
\multicolumn{10}{l}{{\ul \textit{\textbf{Custom Trained Baselines (SFT \& DPO)}}}} \\

\multirow{3}{*}{\textit{\textbf{Hunyuan}}} & 1.8B &
\cellcolor{Blue!60}7.32 ± 0.21 &
\cellcolor{Blue!60}9.51 ± 0.21 &
\cellcolor{Blue!60}7.32 ± 0.21 &
\cellcolor{Blue!70}0.85 ± 0.01 &
\cellcolor{Blue!25}6.97 &
\cellcolor{Blue!25}10.67 &
\cellcolor{Blue!25}8.87 &
\cellcolor{Blue!30}0.88 \\

 & 4B &
\cellcolor{Blue!70}6.57 ± 0.24 &
\cellcolor{Blue!70}8.66 ± 0.35 &
\cellcolor{Blue!70}6.57 ± 0.24 &
\cellcolor{Blue!80}0.87 ± 0.01 &
\cellcolor{Blue!40}6.00 &
\cellcolor{Blue!35}15.67 &
\cellcolor{Blue!40}7.62 &
\cellcolor{Blue!50}0.91 \\

 & 7B &
\cellcolor{Blue!90}{\ul 5.76 ± 0.19} &
\cellcolor{Blue!90}{\ul 7.79 ± 0.53} &
\cellcolor{Blue!90}{\ul 5.76 ± 0.19} &
\cellcolor{Blue!100}\textbf{0.90 ± 0.01} &
\cellcolor{Blue!70}4.82 &
\cellcolor{Blue!70}31.00 &
\cellcolor{Blue!70}6.12 &
\cellcolor{Blue!70}0.93 \\

\hdashline

\multirow{3}{*}{\textit{\textbf{Qwen3}}} & 1.7B &
\cellcolor{Blue!50}7.83 ± 0.12 &
\cellcolor{Blue!50}9.98 ± 0.35 &
\cellcolor{Blue!50}7.83 ± 0.12 &
\cellcolor{Blue!60}0.83 ± 0.01 &
\cellcolor{Blue!20}7.63 &
\cellcolor{Blue!15}6.00 &
\cellcolor{Blue!20}9.66 &
\cellcolor{Blue!25}0.87 \\

 & 4B &
\cellcolor{Blue!80}6.37 ± 0.22 &
\cellcolor{Blue!80}8.51 ± 0.36 &
\cellcolor{Blue!80}6.37 ± 0.22 &
\cellcolor{Blue!90}0.88 ± 0.01 &
\cellcolor{Blue!45}5.80 &
\cellcolor{Blue!45}16.33 &
\cellcolor{Blue!45}7.39 &
\cellcolor{Blue!45}0.91 \\

 & 8B &
\cellcolor{Blue!100}\textbf{5.67 ± 0.25} &
\cellcolor{Blue!100}\textbf{7.69 ± 0.57} &
\cellcolor{Blue!100}\textbf{5.67 ± 0.25} &
\cellcolor{Blue!100}{\ul 0.90 ± 0.02} &
\cellcolor{Blue!90}{\ul 4.56} &
\cellcolor{Blue!80}32.67 &
\cellcolor{Blue!90}{\ul 5.71} &
\cellcolor{Blue!90}{\ul 0.94} \\

\bottomrule
\end{tabular}
}
\caption{\textbf{Main results on materials data quality evaluation.} 
The table reports overall \emph{Accuracy} (Acc) and \emph{Consistency} (Cons) metrics across all evaluated models. For Accuracy, we report the average scores together with the standard error of the mean (SEM). Bold numbers indicate the best overall performance across all models, while underlined numbers denote the best results within each model group.}
\label{tab:main-table}
\end{table}

As reported in Table~\ref{tab:main-table}, all \emph{Custom Trained Baselines} adopt the same two-stage alignment strategy, consisting of SFT cold start followed by DPO with knowledge-augmented preference signals, and use a long-chain output format with \texttt{<think>} during inference. 
Except for entries explicitly marked as ``(with know)'', no external materials knowledge base is accessed during evaluation for any model (rule-free), including our self-trained models. 
This design ensures that evaluation knowledge is internalized through training rather than introduced at inference time, enabling a fair comparison under a rule-free setting.

\paragraph{Accuracy.}
Without external knowledge access, the self-trained models exhibit substantial improvements in accuracy on HEA materials evaluation tasks. 
For example, Hunyuan-7B achieves an $R^2$ of 0.90 on the full test set after training, compared to 0.09 for the untrained base model of the same scale. 
Error metrics decrease consistently, with MAE reduced from 18.77 to 5.76 and RMSE from 23.20 to 7.79. 
Qwen3-8B shows comparable performance, reaching $R^2=0.90$ with MAE 5.67 and RMSE 7.69. In contrast, under the same rule-free setting, the general-purpose large model GPT-5 attains an $R^2$ of 0.60, indicating a clear gap in quantitative alignment when materials evaluation knowledge is not explicitly internalized.

\paragraph{Consistency.}
Beyond accuracy gains, the stability of model outputs across repeated independent evaluations is markedly improved. 
For Hunyuan-7B, the standard deviation of scores decreases from 17.16 to 4.82 (approximately a 72\% reduction), and Pairwise MAD decreases from 21.80 to 6.12. At the same time, Krippendorff’s $\alpha$ increases from 0.59 to 0.93, and Agreement@5 rises from 2.0 to 31.0. 
Qwen3-8B exhibits a consistent improvement trend (Std. 17.65$\rightarrow$4.56, MAD 22.39$\rightarrow$5.71, $\alpha$ 0.59$\rightarrow$0.94, Agreement@5 1.33$\rightarrow$32.67).
These systematic gains indicate a substantial reduction in evaluation uncertainty under identical inputs, thereby enhancing the overall reliability of model judgments.

Across both the Hunyuan and Qwen3 families, increasing model size from 1.8B/1.7B to 4B and further to 7B/8B leads to monotonic or near-monotonic improvements in both Accuracy and Consistency metrics. 
This trend suggests that the proposed two-stage alignment strategy scales stably with model capacity. 
Moreover, compared with general-purpose models equipped with external materials knowledge bases (e.g., ``GPT-5 (with know)'' and ``DeepSeek-R1 (with know)''), our approach achieves comparable consistency metrics (including Std., MAD, and Krippendorff’s $\alpha$) without relying on external knowledge sources, validating the effectiveness of preference-based knowledge internalization.

Overall, under a fully rule-free setting, the self-trained models achieve significant gains across both Accuracy ($R^2$, MAE, RMSE, and Abs Bias) and Consistency (Std., Agreement@5, MAD, and K-$\alpha$). 
Among all evaluated models, Qwen3-8B attains the best overall balance between accuracy and stability, demonstrating that the proposed training framework can simultaneously improve evaluation precision and output consistency. 
These results establish a strong foundation for extending the framework to larger materials systems and cross-task evaluation scenarios.

To further examine whether the models internalize materials evaluation knowledge during training, rather than merely fitting scores in the output space, we construct a multiple-choice knowledge evaluation aligned one-to-one with the SFT-stage QA tasks. 
Focusing on the HEA system, we consider six representative property categories (Phase, Elongation, UTS, HV, Corrosion, and Oxidation). 
For each property, we curate 20 four-option multiple-choice questions that probe explicit knowledge related to phase identification, mechanical property trends, and corrosion or oxidation mechanisms. 
All models directly output answer choices under a rule-free setting, and performance is measured using accuracy (Acc).

\begin{table}[htbp]
\centering
\renewcommand\arraystretch{1.5} 
\resizebox{\textwidth}{!}{
\begin{tabular}{@{}lccccccc@{}}
\toprule
\textbf{Models} & \textbf{Phase} & \textbf{Elongation} & \textbf{UTS} & \textbf{HV} & \textbf{Corrosion} & \textbf{Oxidation} & \textbf{All$^{\bm{\triangle}}$} \\ \midrule
\multicolumn{8}{l}{\textit{\textbf{General Baselines}}} \\
\textbf{gpt-5} & 55.00 ± 11.41 & 75.00 ± 9.93 & 60.00 ± 11.24 & 75.00 ± 9.93 & 70.00 ± 10.51 & 60.00 ± 11.24 & 65.83 ± 4.35 \\
\textbf{DeepSeek-R1} & 60.00 ± 11.24 & 60.00 ± 11.24 & 60.00 ± 11.24 & 65.00 ± 10.94 & 65.00 ± 10.94 & 50.00 ± 11.47 & 60.00 ± 4.49 \\\midrule
\multicolumn{8}{l}{\textit{\textbf{Custom Trained Baselines (SFT \& DPO)}}} \\
\textbf{Hunyuan-7B-Instruct} & 30.00 ± 10.51 & 20.00 ± 9.18 & 35.00 ± 10.94 & 35.00 ± 10.94 & 30.00 ± 10.51 & 20.00 ± 9.18 & 28.33 ± 4.13 \\
\textit{\textbf{+ SFT}} & 70.00 ± 10.51 & 60.00 ± 11.24 & 70.00 ± 10.51 & {\ul 65.00 ± 10.94} & {\ul 80.00 ± 9.18} & {\ul 80.00 ± 9.18} & 70.83 ± 4.17 \\
\textit{\textbf{+ SFT \& DPO}} & \textbf{75.00 ± 9.93} & {\ul \textbf{80.00 ± 9.18}} & {\ul \textbf{95.00 ± 5.00}} & {\ul \textbf{80.00 ± 9.18}} & \textbf{90.00 ± 6.88} & {\ul \textbf{95.00 ± 5.00}} & {\ul 85.83 ± 3.20} \\ \hdashline
\textbf{Qwen3-8B-Instruct} & 40.00 ± 11.24 & 30.00 ± 10.51 & 25.00 ± 9.93 & 30.00 ± 10.51 & 20.00 ± 9.18 & 35.00 ± 10.94 & 30.00 ± 4.20 \\
\textit{\textbf{+ SFT}} & {\ul 80.00 ± 9.18} & {\ul 80.00 ± 9.18} & 70.00 ± 10.51 & 75.00 ± 9.93 & 75.00 ± 9.93 & 70.00 ± 10.51 & 75.00 ± 3.97 \\
\textit{\textbf{+ SFT \& DPO}} & {\ul \textbf{85.00 ± 8.19}} & {\ul \textbf{85.00 ± 8.19}} & {\ul \textbf{90.00 ± 6.88}} & {\ul \textbf{90.00 ± 6.88}} & {\ul 80.00 ± 9.18} & \textbf{95.00 ± 5.00} & \textbf{87.50 ± 3.03} \\ \bottomrule
\end{tabular}%
}
\caption{\textbf{Knowledge-based multiple-choice evaluation on HEAs.}
Accuracy (Acc, \%) of different models on four-option multiple-choice questions designed to probe explicit materials evaluation knowledge across six property categories (Phase, Elongation, UTS, HV, Corrosion, and Oxidation). The questions are aligned with the SFT-stage QA supervision and are evaluated under a rule-free setting without external knowledge access.}
\label{tab:know-table}

\end{table}

As shown in Table~\ref{tab:know-table}, the untrained Hunyuan-7B-Instruct and Qwen3-8B-Instruct models achieve overall accuracies of only 28.33\% and 30.00\%, respectively—close to the theoretical random-guess baseline of 25\% for four-choice questions. 
This result indicates that off-the-shelf instruct models possess little usable structured domain knowledge when applied to materials evaluation tasks.

After introducing SFT cold-start training alone, the overall accuracy of both model families increases substantially, reaching 70.83\% for Hunyuan-7B and 75.00\% for Qwen3-8B. 
Across most property categories, accuracies consistently fall within the 70–80\% range, clearly surpassing general-purpose large models such as GPT-5 (65.83\%) and DeepSeek-R1 (60.00\%). 
These results demonstrate that supervised learning based on materials evaluation rules and explanatory examples is sufficient to establish a structured foundation of domain knowledge within the model.

Building upon SFT, the introduction of DPO-based preference alignment further yields pronounced improvements. 
The overall accuracy increases to 85.83\% for Hunyuan-7B and 87.50\% for Qwen3-8B, with multiple instances of 90–95\% correctness observed for properties such as UTS and Oxidation. 
These results not only exceed the performance of substantially larger models such as GPT-5 and DeepSeek-R1, but also highlight the critical role of \emph{knowledge-augmented preference signals} in correcting biased knowledge representations and unstable cognitive patterns. 
By enforcing preferences between judgments that are both knowledge-correct and reasoning-consistent and those that are superficially plausible yet mechanistically flawed, DPO effectively strengthens correct knowledge representations while suppressing erroneous patterns, without requiring additional large-scale pretraining.

Importantly, these findings indicate that the constructed DPO preference data themselves effectively encode the key knowledge required for materials evaluation. 
By forming learnable contrasts between judgments with explicit knowledge support and those lacking sufficient evidence, preference signals can be stably internalized into model parameters during training, rather than serving merely as external prompts or transient alignment constraints.

Overall, the results of the knowledge-based multiple-choice evaluation form a highly consistent line of evidence with the score-based Accuracy and Consistency metrics reported earlier. 
After two-stage \emph{SFT + DPO} alignment, small open-source models not only achieve substantial gains in end-to-end materials evaluation accuracy and stability, but also demonstrate superior explicit knowledge mastery directly relevant to the task. 
These quantitative results validate the effectiveness of knowledge-augmented preference signals in steering model evaluation behavior from unstable, evidence-poor judgments toward consistent, knowledge-grounded reasoning, and establish DPO-based knowledge injection as a low-cost and effective alignment pathway for materials evaluation scenarios.

\subsection{Ablation Study}
\label{sec:ablation_results}

To further disentangle the contributions of individual components in the proposed framework, we conduct systematic ablation studies on Hunyuan-7B and Qwen3-8B. We evaluate three configurations: an SFT cold start, SFT combined with DPO, and variants with or without long/short reasoning templates (denoted as + SFT \& DPO w/o think; unless otherwise specified, models employ the long-chain reasoning format with \texttt{<think>}). A quantitative comparison of these settings is reported in Table~\ref{tab:ablation-table}.

\begin{table}[htbp]
\centering
\renewcommand\arraystretch{1.45} 
\resizebox{\textwidth}{!}{
\begin{tabular}{@{}lllllcccc@{}}
\toprule
\multicolumn{1}{c}{\multirow{2}{*}{\textbf{Models}}} & \multicolumn{4}{c}{\textbf{Accuracy}} & \multicolumn{4}{c}{\textbf{Consistency}} \\
\cmidrule(l){2-9}
& \multicolumn{1}{c}{\textbf{MAE $\downarrow$}} 
& \multicolumn{1}{c}{\textbf{RMSE $\downarrow$}} 
& \multicolumn{1}{c}{\textbf{Abs Bias $\downarrow$}} 
& \multicolumn{1}{c}{\textbf{$\bm{R^2}$ $\uparrow$}} 
& \multicolumn{1}{c}{\textbf{Std. $\downarrow$}} 
& \multicolumn{1}{c}{\textbf{Agree@5(\%) $\uparrow$}} 
& \multicolumn{1}{c}{\textbf{Pair. MAD $\downarrow$}} 
& \multicolumn{1}{c}{\textbf{Kripp. $\bm{\alpha}$ $\uparrow$}} \\ 
\midrule

\textbf{Hunyuan-7B-Instruct} & 18.77 ± 0.63 & 23.2 ± 0.86 & 18.77 ± 0.63 & 0.09 ± 0.07 & 17.16 & 2 & 21.8 & 0.59 \\
\textbf{\textit{+ SFT}} & 9.13 ± 0.3 & 11.24 ± 0.48 & 9.13 ± 0.3 & 0.79 ± 0.02 & {\ul 4.95} & \textbf{35.67} & 6.13 & {\ul 0.91} \\
\textbf{\textit{+ SFT \& DPO w/o think}} & {\ul 6.16 ± 0.15} & {\ul 8.73 ± 0.2} & {\ul 6.16 ± 0.15} & {\ul 0.87 ± 0.01} & 8.91 & 5.33 & 11.33 & 0.83 \\
\textbf{\textit{+ SFT \& DPO}} & \textbf{5.76 ± 0.19} & \textbf{7.79 ± 0.53} & \textbf{5.76 ± 0.19} & \textbf{0.9 ± 0.01} & \textbf{4.82} & {\ul 31.00} & 6.12 & 0.93 \\ \hdashline
\textbf{Qwen3-8B-Instruct} & 18.26 ± 0.64 & 22.58 ± 1.04 & 18.26 ± 0.64 & 0.13 ± 0.08 & 17.65 & 1.33 & 22.39 & 0.59 \\
\textbf{\textit{+ SFT}} & 8.29 ± 0.18 & 10.47 ± 0.37 & 8.29 ± 0.18 & 0.81 ± 0.01 & {\ul 4.72} & \textbf{40.00} & {\ul 5.85} & {\ul 0.92} \\
\textbf{\textit{+ SFT \& DPO w/o think}} & {\ul 5.96 ± 0.1} & {\ul 8.34 ± 0.48} & {\ul 5.96 ± 0.1} & {\ul 0.88 ± 0.01} & 8.11 & 10 & 10.3 & 0.85 \\
\textbf{\textit{+ SFT \& DPO}} & \textbf{5.67 ± 0.25} & \textbf{7.69 ± 0.57} & \textbf{5.67 ± 0.25} & \textbf{0.9 ± 0.02} & \textbf{4.56} & {\ul 32.67} & \textbf{5.71} & \textbf{0.94} \\
\bottomrule
\end{tabular}
}
\caption{\textbf{Ablation study on the proposed framework.}
Performance comparison of different training configurations on Hunyuan-7B and Qwen3-8B, including \emph{SFT} cold start, \emph{+ SFT + DPO}, and variants with or without long-chain reasoning templates (\texttt{<think>}). Results are reported under a rule-free setting without external knowledge access, and evaluated using both Accuracy and Consistency metrics.}
\label{tab:ablation-table}
\end{table}

\paragraph{Effect of SFT cold start.}
Across both model families, SFT cold start constitutes a critical transition from unconstrained scoring to evidence-aware evaluation. 
Specifically, SFT training reduces MAE from approximately 18 to the 8--9 range, while $R^2$ increases from 0.09 / 0.13 to 0.79 / 0.81 for Hunyuan-7B and Qwen3-8B, respectively. 
Consistency-related metrics, including Std., Pairwise MAD, and Krippendorff’s $\alpha$, also improve substantially. 
These results indicate that rule-driven supervised signals enable the models to acquire basic plausibility constraints over composition--property relationships, thereby suppressing arbitrary or unsupported scoring behavior.

\paragraph{Effect of DPO-based knowledge alignment.}
Introducing DPO on top of SFT further calibrates model behavior from \emph{reasonable scoring} toward \emph{expert-aligned judgment}. 
Rather than refitting scores, DPO leverages the constructed knowledge-augmented preference signals to guide the model’s preference between paired evaluation explanations. 
Under the \emph{+ SFT \& DPO} setting, MAE is further reduced to 5.76 and 5.67 for Hunyuan-7B and Qwen3-8B, respectively, while $R^2$ increases to 0.90. 
At the same time, consistency metrics (Std., Pairwise MAD, and Krippendorff’s $\alpha$) are maintained or slightly improved relative to the SFT-only configuration. 
This observation suggests that DPO does not compromise evaluation stability; instead, it refines the decision boundary toward judgments that better reflect high-quality materials evaluation preferences.

\paragraph{Interaction with reasoning templates.}
Comparisons between long and short reasoning templates further reveal a complementary relationship between DPO and explicit reasoning expression. 
Under the same \emph{SFT + DPO} training regime, removing the \texttt{<think>} template (short reasoning) still yields higher Accuracy than SFT alone, but leads to a noticeable degradation in Consistency metrics. 
In contrast, enabling long-chain reasoning restores Std., Pairwise MAD, and Krippendorff’s $\alpha$ to their best levels while preserving comparable or slightly improved error metrics. 
These results indicate that DPO primarily provides preference-based discriminative signals grounded in domain knowledge, whereas explicit long-form reasoning offers the representational capacity necessary to unfold these signals into stable decision-making. 
Their combination is therefore essential for achieving both high accuracy and high consistency in materials evaluation tasks.

\section*{Discussion}
\label{discussion}
\setcounter{section}{3}

Materials research is increasingly shifting from generating candidates at scale to making robust and reliable judgments. As model-driven candidate generation and high-throughput experimental platforms mature, the main bottleneck in materials discovery is no longer the size of the candidate space, but whether evaluation and screening can remain scalable, reproducible, and consistent. Without stable evaluation mechanisms, ranking decisions become unreliable and resource allocation becomes increasingly uncertain, ultimately limiting the efficiency of closed-loop discovery systems.

To address this challenge, we propose a knowledge-augmented preference alignment framework, \textsc{MaterEval}, that enables small open-source large language models to perform materials evaluation as a core judgment task. Rather than relying on explanation generation or external retrieval at inference time, our approach transforms expert knowledge into learnable preference signals by contrasting evidence-supported judgments with unsupported guesses. Through an automated rule mining and data construction pipeline, combined with a two-stage training strategy based on SFT cold start and DPO alignment, materials evaluation knowledge is stably internalized into model parameters. As a result, the trained models are able to produce consistent and interpretable evaluations under a fully rule-free setting.

Building on this foundation, we further introduce a fast--slow hierarchical evaluation paradigm to balance efficiency and reliability in practical screening scenarios. Fast thinking enables high-throughput preliminary filtering over large candidate pools, while slow thinking supports more rigorous evidence integration and in-depth judgment for a small number of high-potential candidates. Experiments show that this hierarchical design substantially reduces inference cost while preserving evaluation stability and consistency, making the trained evaluators both accurate and practically deployable under candidate explosion settings.

Across multi-objective materials evaluation tasks exemplified by high-entropy alloys, systematic experiments validate the effectiveness of the proposed framework. After training, small models achieve concurrent improvements in both Accuracy and Consistency (e.g., $R^2$ reaching 0.90 and Krippendorff’s $\alpha$ increasing to 0.93--0.94), clearly outperforming untrained base models and matching or exceeding the evaluation stability of some general-purpose large models without accessing external knowledge bases. Independent knowledge-based evaluations further show that the models acquire strong task-relevant domain knowledge, indicating that preference-based training improves both scoring behavior and underlying knowledge representations.

\vspace{0.2cm}
\noindent \textbf{Future Directions:}

\begin{enumerate}
\item \textbf{Closed-loop validation and online calibration.} 
Future work will explore integrating the evaluation models with real experimental feedback or high-fidelity simulations, enabling continuous calibration of preference signals and judgment boundaries within a closed-loop pipeline of \emph{generation--evaluation--validation--update}. Such a setting would allow model judgments to better align with practical synthesizability and observed experimental outcomes.

\item \textbf{Hierarchical preference signals and curriculum-style alignment.} 
Another promising direction is to introduce hierarchical or difficulty-aware preference construction strategies, allowing models to progress from simple criteria to multi-evidence trade-offs and high-uncertainty judgments. This curriculum-style alignment may further improve training stability and efficiency in complex evaluation scenarios.
\end{enumerate}

Overall, this work shows that improving evaluation capability in autonomous materials discovery does not require larger models or more complex external tools. 
By transforming domain knowledge into preference signals and aligning model behavior through structured training and hierarchical reasoning, small open-source LLMs can deliver stable, reliable, and deployable materials evaluation. 
More broadly, these findings suggest that evaluation-centric alignment provides a general and scalable pathway for integrating language models into scientific decision-making workflows.

\section*{Methods}
\label{sec:methods}
\setcounter{section}{4}
\setcounter{subsection}{0}

\subsection{Knowledge-Augmented Preference Dataset Construction}
\label{sec:dataset}

In closed-loop autonomous materials discovery, screening and evaluation often become throughput bottlenecks earlier than candidate generation. 
While candidate materials can be generated at scale through model inference and high-throughput pipelines, experimental success ultimately depends on the ability to evaluate candidates in a low-cost, scalable, and reproducible way—namely, to determine whether they are trustworthy, synthesizable, and worth prioritizing for validation.
Such evaluation is not a simple text generation or property regression task. 
Instead, it constitutes a highly knowledge-intensive \emph{judgment} problem, requiring models to invoke coupled composition–processing–property knowledge, follow verifiable criteria chains, and produce stable, consistent conclusions accompanied by interpretable risk explanations.

However, directly applying general-purpose LLMs to materials evaluation often leads to superficially plausible conclusions that lack traceable evidence, with judgments that drift across different prompts or formulations. 
Retrieval-augmented approaches based on large closed-source models can improve single-instance evaluation quality, but they are difficult to scale when candidate volumes grow rapidly, due to prohibitive cost and limited throughput. 
These limitations raise a central question: how can dispersed expert knowledge be transformed into learnable and scalable supervision signals, so that small open-source LLMs can acquire effective evaluation capabilities in a fully rule-free setting and learn to make evidence-grounded judgments?

\begin{figure}
	\centering
	\includegraphics[width=1\textwidth]{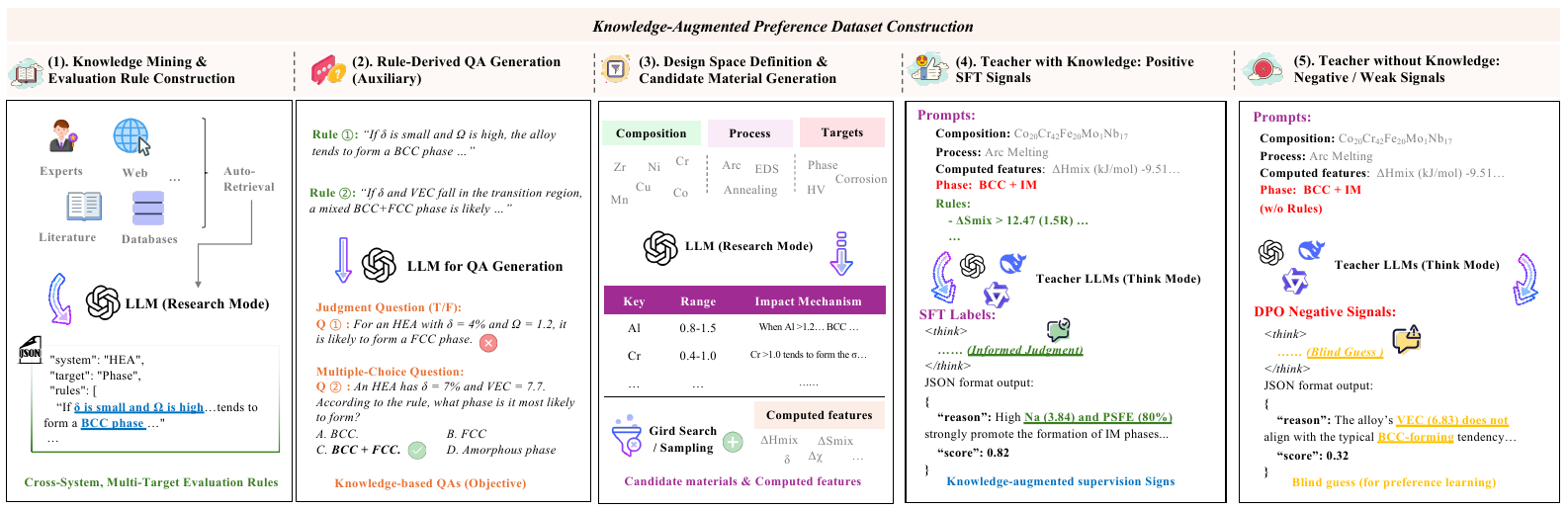}
	\caption{\textbf{Knowledge-augmented preference dataset construction pipeline for material evaluation.} The pipeline consists of five stages: (1) mining domain knowledge from literature and databases to construct cross-system, multi-target evaluation rules; (2) generating rule-derived objective QAs (T/F and multiple-choice) as auxiliary supervision; (3) defining the design space with a unified schema (composition–process–targets) and computing key material descriptors; (4) producing knowledge-grounded judgments as positive/strong SFT signals using a teacher with explicit rules; and (5) generating rule-free “blind-guess” judgments as negative/weak signals for preference alignment.}
	\label{fig:dataset}
\end{figure}

To address the above challenges, we design and implement a \emph{Knowledge-Augmented Preference Dataset Construction} pipeline (Fig.~\ref{fig:dataset}). 
Rather than reducing evaluation capability to explanation generation, the core idea is to explicitly construct preference signals that distinguish \emph{informed judgments} from \emph{blind guesses}. 
This enables models to internalize how to reason with evidence and to avoid improper knowledge usage and unstable conclusions during training. 
As illustrated in Fig.~\ref{fig:dataset}, the pipeline consists of five fully automated and low-cost components, allowing scalable data generation for multi-objective materials evaluation.

\paragraph{(1) Knowledge mining and evaluation rule construction.}
We automatically retrieve and distill actionable domain knowledge for multi-objective materials evaluation from the literature, professional databases, and web resources, and abstract implicit expert experience into a structured rule library. 
The rules capture not only conditions likely to hold, but also signals of potential risk or unreliability, forming reusable criteria across material systems (e.g., valid ranges of key descriptors, their mechanistic links to phase or properties, and common violations of process or mechanism consistency). This step converts experts’ implicit scrutiny and skepticism into explicit evidence units for downstream training.

\paragraph{(2) Rule-derived QA generation (auxiliary).}
Open-ended evaluation alone may encourage surface-level plausibility without robust criterion grounding. 
To mitigate this, we automatically derive objective auxiliary QA tasks from the rule library, including true/false and multiple-choice questions. 
These QAs serve as cold-start supervision to align models with core criteria and boundary conditions, providing a grounded foundation for subsequent holistic evaluation.

\paragraph{(3) Design space definition and candidate generation.}
To ensure a consistent evaluation context, we adopt a unified input schema that binds \emph{Composition–Process–Properties} as the basic evaluation unit. 
For each candidate, we compute commonly used materials descriptors (e.g., $\Delta H_{\mathrm{mix}}$, $\Delta S_{\mathrm{mix}}$, $\Delta\chi$, VEC, $\delta$) \citep{zeng2021revealing, singh2023phase} and align them with thresholds and mechanistic entries in the rule library. 
This design provides checkable evidence sources alongside conclusions, reducing hallucinations driven purely by linguistic plausibility.

\paragraph{(4) Teacher with knowledge: positive (strong) SFT signals.}
To generate high-quality supervision, we explicitly provide teachers with rules and evidence constraints and require outputs to follow a predefined JSON schema, yielding a complete evidence-to-conclusion chain. 
Each output includes (i) a \texttt{<think>} reasoning trace that shows how evidence is invoked and weighted, and (ii) a structured summary that compresses and organizes the evidence. The schema enables direct parsing of quantitative results (e.g., labels or scores) together with reusable evidence summaries. 
These samples form the positive SFT signals corresponding to \emph{informed judgments}, guiding models toward reproducible and interpretable evaluations.

\paragraph{(5) Teacher without knowledge: negative (weak) signals.}
To construct learnable preference contrasts, we remove rules and evidence constraints under the same candidate inputs (w/o rules), prompting teachers to produce judgments that are more drift-prone and reliant on linguistic priors. 
These outputs are not treated as incorrect answers; instead, they serve as negative signals corresponding to \emph{blind guesses}. 
By pairing knowledge-grounded and knowledge-absent outputs, preference alignment systematically discourages evidence-free, mechanistically inconsistent, or unstable evaluation paths—even when explanations appear fluent—thereby shifting model capability from \emph{explaining} to \emph{judging}.

This dataset construction pipeline is designed to provide a scalable supervision mechanism for materials evaluation, with a formulation that is decoupled from any specific materials system. 
Centered on a unified \emph{composition–process–target property} input paradigm, the pipeline combines structured domain knowledge (e.g., evaluation rules and criteria distilled from the literature and databases) with computable or extractable materials descriptors to automatically generate training samples suitable for both SFT and preference alignment.

We select high-entropy alloys (HEAs) as a representative materials system. HEAs are characterized by multi-principal-element compositions, whose design space grows combinatorially with elemental combinations \citep{yu2024recent, yin2025composition}. Moreover, their phase stability and synthesizability strongly depend on coupled composition–process relationships, leading to high knowledge density and uncertainty in evaluation tasks. 
These properties make HEAs a suitable testbed for systematically assessing the effectiveness and consistency of the constructed supervision signals.

Under the HEA setting, we construct datasets for six evaluation objectives: phase constitution (Phase), elongation (Elongation), ultimate tensile strength (UTS), microhardness (HV), corrosion resistance (Corrosion), and oxidation resistance (Oxidation). The primary evaluation dataset contains 47,472 samples (7,912 per objective), accompanied by 1,732 auxiliary knowledge-based QA samples. 
The preference learning dataset for DPO \citep{rafailov2023direct} covers the same six objectives and comprises 6,000 samples in total (1,000 per objective). The DPO data are formed by combining two sources: (i) a subset of high-loss samples from the SFT stage, which target evaluation cases where the model remains insufficiently aligned, and (ii) newly sampled candidates from the design space to increase distributional coverage and enhance preference signal diversity.

\subsection{Two-stage Alignment Training: From Blind Guess to Informed Judgment}
\label{sec:training}

We adopt a two-stage alignment training strategy to shift the behavior of small open-source LLMs in materials evaluation from \emph{blind guesses} to \emph{informed judgments} (Fig.~\ref{fig:framework}, left). 
A key design choice is that the student model does not explicitly receive rule texts during either training or inference (rule-free). 
The model takes as input only structured information $x$ describing the candidate material, including composition, processing conditions, computable features, and the evaluation target. 
Domain knowledge is not injected via prompts; instead, it is internalized into model parameters through constructed supervision signals. 
Positive samples encode how judgments should be formed based on evidence, while negative samples capture common drift and instability when evidence is absent. 
This design enables the model to acquire scalable and deployable materials evaluation capabilities at controlled cost.

\paragraph{Input representation.}

As illustrated in Fig.~\ref{fig:framework}, each evaluation instance is organized into a unified input $x$ comprising \emph{Composition}, \emph{Process}, and a set of target-relevant computed features (e.g., $\Delta H_{\mathrm{mix}}$, $\Delta S_{\mathrm{mix}}$, VEC, $\delta$), with the evaluation objective explicitly specified (e.g., Phase, UTS, HV). 
This input format is kept identical during training and testing, preventing reliance on rule prompts and better reflecting practical deployment scenarios for large-scale candidate screening.

\paragraph{Phase I: SFT cold start (rule-free learning with structured supervision).}

In the first stage, we perform SFT to cold-start the model, enabling it to produce interpretable and evidence-grounded evaluation outputs and to acquire basic judgment chains for materials evaluation. 
Specifically, we decompose the positive supervision signal into two complementary output forms,
$
y^{+}=\{y^{+}_{l},y^{+}_{s}\},
$
to support different granularity and usage scenarios:
\begin{enumerate}
	\item \emph{Long reasoning output} $y^{+}_{l}$, which presents more complete evidence organization and reasoning trajectories, corresponding to the Long Reasoning Data in Fig.~\ref{fig:framework};
	\item \emph{Short judgment output} $y^{+}_{s}$, which provides concise conclusions suitable for high-throughput screening scenarios, corresponding to the Short Judgment Data.
\end{enumerate}

To bind evaluation conclusions with their supporting evidence into supervision signals that can be automatically processed, we enforce a unified JSON schema on teacher outputs. Each output includes: (i) a reasoning trace enclosed in the \texttt{<think>} tag, describing how evidence is invoked and weighed; (ii) a summarized rationale that compresses the reasoning for efficient storage and alignment; and (iii) directly parsable quantitative evaluation results (e.g., a \texttt{score} field). 
This design preserves rich judgment trajectories while enabling stable extraction of the core supervision signal, namely \emph{quantitative outcome + rationale summary}, for subsequent analysis and evaluation.

In addition, we incorporate rule-derived objective QA tasks (true/false and multiple-choice questions) as auxiliary cold-start supervision $(x_Q, y_Q)$ to strengthen the model’s alignment with key evaluation criteria and conceptual boundaries (the QAs module in Fig.~3). 
The overall training objective is defined as
$$
\begin{gathered}
\mathcal{L}_{\mathrm{SFT}}=\mathbb{E}_x\left[-\log \pi_\theta\left(y^{+} \mid x\right)\right]
+\mathbb{E}_{x_Q}\left[-\log \pi_\theta\left(y_Q \mid x_Q\right)\right], \\
\text{where } y^{+}=\left\{y^{+}_{l}, y^{+}_{s}\right\}.
\end{gathered}
$$

\noindent where $\pi_\theta$ denotes the student model with parameters $\theta$; $x$ is the structured evaluation input (composition, process, computed features, and target); $y^{+}$ represents positive evaluation outputs; $(x_Q, y_Q)$ denotes auxiliary QA inputs and labels derived from rules.

This stage enables the model, without explicitly reading rule texts, to generate structured judgments solely from input features and to consistently exhibit a parsable \emph{conclusion–evidence–summary} output pattern.

\paragraph{Phase II: DPO knowledge alignment (preference learning against blind guess).}
In the second stage, we apply preference alignment to further strengthen the robustness and consistency of model judgments. For the same input $x$, we construct paired preference samples:
\begin{itemize}
	\item $y^{+}$, an \emph{informed judgment} generated by a knowledge-conditioned teacher with rule and evidence support; 
	\item $y^{-}$, a \emph{blind guess} produced after removing rule support, which is more prone to mechanistic inconsistency, insufficient evidence, or conclusion drift.
\end{itemize}

We adopt the DPO~\citep{rafailov2023direct} objective to explicitly model the model’s relative preference between positive and negative samples, encouraging the parameterized model to favor $y^{+}$ while suppressing behaviors associated with $y^{-}$:

$$
\mathcal{L}_{\mathrm{DPO}}=-\log \sigma\left(\beta\left[\log \pi_\theta\left(y^{+} \mid x\right)-\log \pi_\theta\left(y^{-} \mid x\right)\right]\right)
$$
\noindent where $\sigma(\cdot)$ is the sigmoid function; and $\beta$ is a temperature parameter controlling preference sharpness.

Importantly, the objective of this stage is not to promote more verbose or expressive explanations. 
Instead, DPO shifts the generation distribution toward evaluation paths that are evidence-grounded, mechanistically consistent, and decision-stable, while demoting judgments that rely primarily on linguistic plausibility without sufficient support.

Following the proposed training loop, the DPO dataset is constructed by combining two complementary sources rather than resampling independently from scratch. 
First, we include a subset of \emph{high-loss} samples from the SFT stage (measured by negative log-likelihood), which target evaluation cases where the model remains insufficiently aligned. 
Second, we incorporate newly sampled instances from the candidate space to expand distributional coverage and enhance preference signal diversity. 
This composition allows preference alignment to correct model weaknesses while maintaining generalization to newly encountered evaluation scenarios.

\subsection{Automated Evaluation: Joint Assessment of Accuracy and Consistency}
\label{sec:evaluation}

Materials evaluation requires not only accurate judgments (\emph{Accuracy}) but also stable and reproducible outputs (\emph{Consistency}). 
In large-scale candidate screening, instability in evaluation can directly distort ranking structures, leading to inefficient resource allocation and increased decision noise. 
We therefore adopt an automated evaluation protocol on an independent test set to jointly quantify Accuracy and Consistency (Fig.~\ref{fig:framework}, right). 
Consistent with the training setup, all evaluated models receive only rule-free structured inputs $x$; evaluation rules and reference scores are used solely to construct computable benchmarks and are not provided as prompts during inference.

\paragraph{Independent test set and ground-truth aggregation.}
To obtain reproducible and relatively robust reference values, we employ a rule-consistent evaluation strategy with multiple strong teacher models on the independent test set. 
Under identical evaluation rules and unified input representations, each teacher independently evaluates every sample multiple times (three repetitions per teacher), producing a set of reference scores. 
These scores are then aggregated across teachers and repetitions by averaging, yielding the reference ground truth for each evaluation objective.

This \emph{multiple-teachers $\times$ repeated evaluations} aggregation serves two purposes: (i) it reduces the influence of biases from any single model, and (ii) it averages out unavoidable stochasticity and output variability in evaluation. 
As a result, it provides a more stable and reproducible reference baseline for the joint assessment of Accuracy and Consistency.

\paragraph{Multi-sample probing and score parsing.}
To assess output stability, each test sample is evaluated through multiple independent generations, yielding a sequence of scores $\{s^{(1)}, \ldots, s^{(T)}\}$. 
As model outputs follow a unified JSON schema, quantitative results can be reliably and automatically parsed from the \texttt{score} field, enabling downstream statistical analysis. 
Based on these parsed scores, we compute two complementary metrics:

\emph{Accuracy.} We measure the deviation between the predicted mean
$\bar{s}=\frac{1}{T}\sum_{t=1}^{T} s^{(t)}$
and the ground-truth mean $\mu$, aggregated across the test set into metrics such as MAE or absolute bias. 
This corresponds to the alignment from $\mu$ to $\bar{s}$ in Fig.~\ref{fig:framework} (``\ding{172} Accuracy'').

\emph{Consistency.} We characterize the dispersion of the score sequence $\{s^{(t)}\}$, using statistics such as the standard deviation or robust pairwise difference measures, to quantify the stability of model judgments under stochastic sampling (Fig.~\ref{fig:framework}, ``\ding{173} Consistency'').

This evaluation protocol explicitly decouples \emph{closeness to the reference standard} (Accuracy) from \emph{output stability under repeated calls} (Consistency). 
The former assesses whether a model produces quantitatively correct judgments, while the latter captures whether these judgments remain stable across independent generations—both of which are critical for reliable ranking and decision-making in practical materials screening.

\paragraph{QA Accuracy (knowledge probing).}
Beyond score-based Accuracy and Consistency, we introduce a third evaluation dimension, \emph{QA Accuracy} (Fig.~\ref{fig:framework}, ``\ding{174} QAs''), to examine whether models internalize explicit materials evaluation knowledge rather than merely fitting scores in a numerical space. 
We construct an independent set of objectively verifiable knowledge questions, including true/false and multiple-choice items, covering core evaluation criteria such as phase stability conditions, mechanical property trends, and mechanisms and risk factors related to corrosion and oxidation.

During testing, models directly output answers or option selections under the same rule-free setting, and performance is measured by classification accuracy. 
QA Accuracy complements score-based metrics: while QA Accuracy probes explicit mastery of evaluation rules and domain knowledge, score-based Accuracy and Consistency assess quantitative alignment and stability in end-to-end judgment tasks. 
Together, these metrics provide a more comprehensive characterization of a model’s materials evaluation capability.

% \backmatter

% \bmhead{Supplementary information}

% If your article has accompanying supplementary file/s please state so here. 

% Authors reporting data from electrophoretic gels and blots should supply the full unprocessed scans for key as part of their Supplementary information. This may be requested by the editorial team/s if it is missing.

% Please refer to Journal-level guidance for any specific requirements.

\section*{Data availability}

All datasets used in this study are publicly available. The knowledge-augmented preference datasets constructed for materials evaluation, including SFT, DPO, and auxiliary QA data, are released on Hugging Face at  
\url{https://huggingface.co/datasets/yuyouyu/MaterEvalData}.

In addition, the trained model weights for the proposed framework, including the Qwen3-8B model aligned with the MaterEval training pipeline, are available at  
\url{https://huggingface.co/yuyouyu/Qwen3-8B-MaterEval}.

\section*{Code availability}

The source code for dataset construction and automated evaluation is publicly available at  
\url{https://github.com/yuyouyu32/MaterEval}.

This repository includes implementations for knowledge mining, preference dataset generation, and the evaluation protocols for accuracy, consistency, and QA-based knowledge probing. Model training for supervised fine-tuning (SFT) and direct preference optimization (DPO) was conducted using the LLamaFactory~\citep{zheng-etal-2024-llamafactory} framework, with configuration files and training scripts adapted accordingly.

\begin{appendices}

\renewcommand{\thefigure}{S\arabic{figure}}
\renewcommand{\thetable}{S\arabic{table}}
\setcounter{figure}{0}
\setcounter{table}{0}




\section{Implementation Details}
\label{sec:implementation}

\subsection{Training Configuration}

All models are trained using a unified two-stage alignment strategy consisting of supervised fine-tuning (SFT) followed by direct preference optimization (DPO). Training is conducted under a fully rule-free setting, where the model inputs contain only structured material descriptors and task specifications.

Unless otherwise stated, the following training hyperparameters are used for both stages:
\begin{itemize}
    \item \textbf{Per-device batch size}: 4
    \item \textbf{Gradient accumulation steps}: 4 (effective batch size = 128 tokens per update)
    \item \textbf{Learning rate}: $1.0 \times 10^{-4}$
    \item \textbf{Number of epochs}: 3
    \item \textbf{Learning rate scheduler}: cosine decay
    \item \textbf{Warmup ratio}: 0.1
    \item \textbf{Precision}: bfloat16 (bf16)
\end{itemize}

We adopt the AdamW optimizer with default momentum parameters ($\beta_1=0.9, \beta_2=0.999$) and weight decay set to $0.01$. Gradient clipping with a maximum norm of $1.0$ is applied to stabilize optimization. All models are trained with fully sharded data parallelism using DeepSpeed ZeRO-2 to reduce memory overhead and improve training efficiency.

For DPO training, preference pairs are constructed on-the-fly from the pre-generated dataset. 
The preference strength coefficient $\beta$ is fixed across experiments to ensure consistent alignment behavior, and no additional reward models or auxiliary critics are introduced.

Unless otherwise specified, DPO training uses the following hyperparameters:
\begin{itemize}
    \item \textbf{Per-device batch size}: 2
    \item \textbf{Gradient accumulation steps}: 8
    \item \textbf{Learning rate}: $5.0 \times 10^{-6}$
    \item \textbf{Number of epochs}: 3
    \item \textbf{Learning rate scheduler}: cosine decay
    \item \textbf{Warmup ratio}: 0.1
    \item \textbf{Precision}: bfloat16 (bf16)
\end{itemize}

We adopt the standard DPO objective with a sigmoid-based preference loss. 
The preference scaling factor is set to $\beta = 0.1$ for all experiments.

\subsection{Inference Settings}

During inference, all LLMs are evaluated under identical generation configurations to ensure fair comparison. We use standard decoding parameters commonly adopted in LLM evaluation:
\begin{itemize}
    \item \textbf{Sampling temperature}: 0.7, controlling the diversity of generated judgments
    \item \textbf{Top-$p$ (nucleus sampling)}: 0.95, ensuring coverage of 95\% cumulative probability mass
    \item \textbf{Maximum generation length}: 4096 tokens
\end{itemize}

For consistency evaluation, each test instance is independently sampled multiple times under the same decoding configuration, with different random seeds. All model outputs follow a unified JSON schema, allowing stable parsing of quantitative scores and structured rationales without post-hoc heuristics.

\subsection{Evaluation Protocol}

Evaluation is conducted on an independent test set that is strictly disjoint from all training data. The assessment protocol jointly measures accuracy and consistency, reflecting both the correctness and stability of model judgments.

For each test sample, multiple independent generations are produced. Quantitative scores are parsed from the structured output and aggregated as follows:
\begin{itemize}
    \item \textbf{Accuracy}: measured by the deviation between the mean predicted score and the aggregated ground-truth reference, reported using MAE, RMSE, Abs Bias, and $R^2$
    \item \textbf{Consistency}: quantified by the dispersion across repeated generations, including Std., Pairwise MAD, Agreement@5, and Krippendorff’s $\alpha$
\end{itemize}

In addition, explicit knowledge internalization is evaluated via a separate set of rule-derived objective QAs (T/F and multiple-choice), where models directly output answers under the same rule-free inference setting. Accuracy on these QAs provides a complementary probe of domain knowledge acquisition beyond score-level fitting.

\subsection{Hardware and Software Environment}

All experiments are conducted on a single-node server equipped with 8 NVIDIA H20 GPUs. The software environment is configured as follows:
\begin{itemize}
    \item \textbf{Python}: 3.11.2
    \item \textbf{PyTorch}: 2.6.0
    \item \textbf{Transformers}: 4.50.0
    \item \textbf{Accelerate}: 1.2.1
    \item \textbf{DeepSpeed}: 0.16.4
    \item \textbf{CUDA}: 12.2
\end{itemize}

All training and evaluation pipelines are fully deterministic given fixed random seeds, enabling reproducibility across runs.

\section{Prompt Templates}
\label{sec:prompts}

To construct high-quality supervision signals for materials evaluation, we design a structured prompt to guide a large teacher model in performing knowledge-grounded judgment over model-predicted HEA states. As shown in Table~\ref{tab:data_generate_prompt}, the prompt explicitly frames the task as a materials evaluation problem, requiring the teacher model to assess the plausibility and reliability of predicted material outcomes by jointly considering alloy composition, processing conditions, calculated thermodynamic descriptors, empirical judgment rules, and available evidence from similar experimentally reported HEAs.

Rather than directly predicting material properties or experimental outcomes, the prompt encourages the teacher model to emulate expert-level evaluation behavior, integrating multiple sources of domain knowledge into a coherent judgment. The teacher model is instructed to provide both a structured numerical score and an accompanying rationale, enabling the resulting annotations to capture not only final assessment outcomes but also the underlying evaluative logic. These structured outputs serve as the primary supervision signals for subsequent supervised fine-tuning and preference alignment, allowing smaller open-source language models to internalize materials evaluation criteria without relying on external retrieval or rule-based systems at inference time.

\begin{table}[htbp]
\centering
\renewcommand\arraystretch{1}
\resizebox{\textwidth}{!}{
\begin{tabular}{@{}l@{}}
\toprule
\textbf{Prompt for Teacher-Model–Driven Materials Evaluation in HEAs} \\
\hline
You are a materials science expert specializing in High Entropy Alloys (HEAs), with expertise in \{target\_description\}. \\
\\
Your task is to evaluate whether a given HEA has \textbf{experimental validation potential}, i.e., the likelihood that the alloy \\
can form the model-predicted \{target\} phase or structure in real experimental conditions. \\
\\
The evaluation should be based on the following information: \\
-- Alloy composition, processing conditions, and calculated descriptors (e.g., mixing entropy, mixing enthalpy); \\
-- The \{target\} structure predicted by the model; \\
-- Similar known HEAs with experimental evidence (may be empty); \\
-- Experiment-related judgment rules (RULES). \\
\\
Provide a detailed reasoning process to ensure scientific reliability and consistency: \\
1. Examine whether the predicted \{target\} is consistent with empirical knowledge given the composition and descriptors; \\
2. Apply the judgment rules to assess the theoretical feasibility of forming the predicted \{target\}; \\
3. Compare with similar known HEAs and evaluate whether experimental evidence supports the prediction (optional); \\
4. Integrate rule compliance and experimental support to assign a potential score; \\
5. Clearly justify the assigned score based on the above analysis. \\
\\
\textbf{RULES:} \\
\{Rules\} \\
\\
\textbf{Similar Real HEAs:} \\
\{sim\_HEAs\} \\
\\
\textbf{DATA:} \\
Composition: \{composition\} \\
Processing conditions: \{process\_desc\} \\
Calculated descriptors: \{calculate\_desc\} \\
Model-predicted \{target\}: \{predicted\_target\_value\} \\
\\
The experimental validation potential score should range from 0 to 1, where 1 indicates high likelihood of experimental \\
realizability and strong consistency with HEA knowledge, and 0 indicates very low feasibility. \\
\\
Output the final evaluation result in the following JSON format: \\
\{ \\
\qquad ``score": Experimental validation potential score in \lbrack0, 1\rbrack, rounded to two decimal places, \\
\qquad ``reason": Brief explanation of the scoring rationale \\
\} \\
\\
Please note that the complete reasoning process must be provided before the final JSON output to ensure correctness \\
and interpretability of the evaluation. \\
\bottomrule
\end{tabular}
}
\caption{Prompt template used to instruct a large teacher model to perform knowledge-grounded materials evaluation for HEAs.}
\label{tab:data_generate_prompt}
\end{table}

Table~\ref{tab:training_eval_prompt} summarizes the system--user prompt template used to train language models for materials evaluation in HEAs. The system prompt establishes the expert role and domain context, while the user prompt formulates the evaluation task, requiring the model to assess the plausibility of model-predicted material states based on alloy composition, processing conditions, and calculated descriptors, and to output a numerical score with a brief justification.

The prompt template supports two evaluation modes through an optional \texttt{think\_prompt}: a \emph{w/o-think} variant that elicits concise judgments for fast, high-throughput screening, and a \emph{think} variant that encourages more detailed evaluative reasoning for careful assessment of shortlisted candidates. These two modes correspond to the fast--slow evaluation paradigm and are instantiated from the same underlying prompt structure. Importantly, the same prompt template is consistently used during both training and inference, with evaluation depth controlled by a lightweight prompt switch, ensuring alignment between learned evaluation behavior and deployment-time usage.

\begin{table}[htbp]
\centering
\renewcommand\arraystretch{1}
\resizebox{\textwidth}{!}{
\begin{tabular}{@{}l@{}}
\toprule
\textbf{Training and Inference Prompt for Materials Evaluation in HEAs} \\
\hline
\textbf{System Prompt:} \\
You are a materials science expert specializing in High Entropy Alloys (HEAs), with expertise in \{target\_description\}. \\
\\
\textbf{User Prompt:} \\
Your task is to evaluate whether a given alloy exhibits \textbf{experimental validation potential}, defined as the \\
likelihood that the alloy can form the model-predicted \{target\} phase or structure under real experimental conditions. \\
Based on your analysis, assign a score between 0 and 1 and provide a clear justification for the score. \\
\\
The basic information of the alloy is as follows: \\
-- Composition: \{composition\} \\
-- Processing conditions: \{process\_desc\} \\
-- Calculated descriptors: \{calculate\_desc\} \\
-- Model-predicted \{target\}: \{predicted\_target\_value\} \\
\\
\{think\_prompt\} \\
\bottomrule
\end{tabular}
}
\caption{System and user prompt template used during training and inference to LLMs for materials evaluation in HEAs.}
\label{tab:training_eval_prompt}
\end{table}

In addition to judgment-oriented evaluation prompts, we design a dedicated prompt template to construct target-related knowledge question--answer (QA) data using a large teacher model. As shown in Table~\ref{tab:knowledge_qa_prompt}, this prompt guides the teacher model to generate scientifically grounded QAs based on established knowledge of High Entropy Alloys (HEAs), focusing on a specific target property or structure.

The prompt explicitly conditions QA generation on a provided target-related knowledge context, ensuring that the resulting questions and answers reflect accepted physical mechanisms, empirical trends, and judgment principles rather than instance-specific observations. 
These teacher-generated QAs are directly used as auxiliary training data, serving as explicit supervision signals that inject target-related domain knowledge into the model. By providing structured question--answer pairs grounded in established HEA principles, the QAs facilitate knowledge internalization and cold-start alignment, complementing instance-level evaluation data used in subsequent training stages for materials evaluation tasks.

By decoupling explicit knowledge acquisition from instance-level judgment, this design complements the evaluation-oriented prompts and contributes to more stable and interpretable learning behavior.

\begin{table}[htbp]
\centering
\renewcommand\arraystretch{1}
\resizebox{\textwidth}{!}{
\begin{tabular}{@{}l@{}}
\toprule
\textbf{Prompt for Teacher-Model–Driven Knowledge QA Generation in HEAs} \\
\hline
You are a materials science expert in the field of High Entropy Alloys (HEAs). \\
\\
Based on the current scientific understanding of HEAs, generate a set of high-quality question--answer (QA) pairs related to the \\
target property or structure \textbf{\{target\}}. The generated QAs should assess core domain knowledge relevant to \{target\}, \\
including physical mechanisms, empirical trends, and commonly accepted judgment principles. \\
\\
The QAs should be grounded in the following target-related knowledge context: \\
\textbf{\{target\_knowledge\}} \\
\\
Each QA pair should be self-contained, scientifically accurate, and independent of any specific alloy instance, such that it can be \\
used as auxiliary supervision for training and evaluating language models in materials science tasks. \\
\\
\textbf{Output Format:} \\
Please output the generated QAs in a structured JSON format, where each item contains a \texttt{question} field and a corresponding \\
\texttt{answer} field. \\
\bottomrule
\end{tabular}
}
\caption{Prompt template used to instruct a large teacher model to generate target-related knowledge question--answer pairs for HEAs.}
\label{tab:knowledge_qa_prompt}
\end{table}

\section{Case Study}
\label{sec:case_study}

While quantitative metrics such as accuracy and consistency provide a global assessment of model behavior, they do not fully reveal \emph{how} different models arrive at their judgments or \emph{why} their conclusions diverge in specific material scenarios. 
To complement the large-scale statistical evaluation, we therefore present some representative case studies that examines the qualitative reasoning patterns of different models when evaluating the same candidate material under an identical, rule-free input setting.

\begin{figure}
    \centering
    \includegraphics[width=1\textwidth]{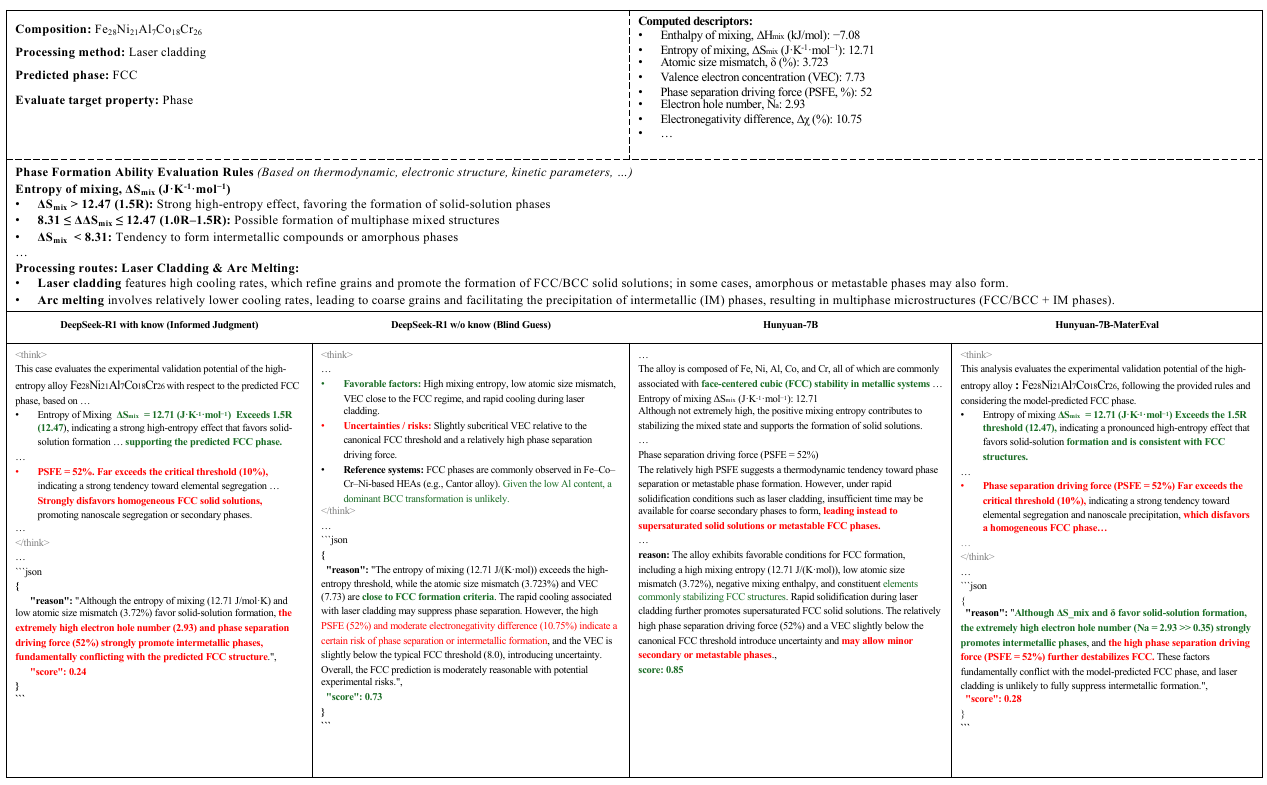}
    \caption{
Qualitative comparison of material phase evaluation for a single HEA case.
The figure shows the evaluation results produced by four models for the same candidate alloy Fe$_{28}$Ni$_{21}$Al$_7$Co$_{18}$Cr$_{26}$ fabricated via laser cladding, with the predicted phase being FCC.
DeepSeek-R1 with explicit knowledge support (\emph{with know}) integrates both stabilizing and destabilizing evidence and assigns a low confidence score, reflecting an informed judgment.
In contrast, the same model without knowledge support (\emph{w/o know}) relies primarily on partial heuristics and assigns a substantially higher score.
A similar discrepancy is observed between the unaligned Hunyuan-7B model and the aligned Hunyuan-7B-MaterEval model, where preference-aligned training leads to structured, evidence-grounded evaluation and a markedly lower confidence score.
Ellipses indicate omitted reasoning details for clarity.
}
    \label{fig:phase_case_study}
\end{figure}

Fig.~\ref{fig:phase_case_study} compares the evaluation results produced by four models for a single high-entropy alloy, Fe$_{28}$Ni$_{21}$Al$_7$Co$_{18}$Cr$_{26}$, fabricated via laser cladding, with the predicted phase being FCC. 
The task is to assess the experimental validation potential of this phase prediction.

A clear distinction can be observed between \emph{informed judgment} and \emph{blind guess}. 
When explicit domain knowledge is available, DeepSeek-R1 (\emph{with know}) integrates both stabilizing and destabilizing factors, explicitly weighing high mixing entropy and low atomic size mismatch against strong phase separation tendencies (PSFE = 52\%) and an elevated electron hole number. 
By accounting for these conflicting signals, the model identifies a fundamental inconsistency with the predicted FCC phase and assigns a low confidence score (0.24). 
In contrast, the same model operating without knowledge support (\emph{w/o know}) relies primarily on partial heuristics—such as entropy arguments, approximate VEC criteria, and analogies to canonical FCC HEAs—while downplaying strong destabilizing indicators, leading to a substantially higher score (0.73) despite acknowledging similar uncertainties.

A comparable discrepancy is observed for the open-source models. 
The unaligned Hunyuan-7B model produces a high confidence score (0.85), largely driven by generic FCC-stabilizing arguments and rapid solidification effects, without consistently resolving conflicting evidence. 
Moreover, its output does not reliably follow the required structured JSON format, suggesting that, without alignment, the model treats material evaluation as free-form explanation rather than a disciplined judgment task. 
In contrast, the aligned Hunyuan-7B-MaterEval model explicitly emphasizes adverse indicators such as the high phase separation driving force and electron hole number, assigns a low score (0.28), and produces a stable, structured judgment consistent with the informed teacher.

\begin{figure}
    \centering
    \includegraphics[width=1\textwidth]{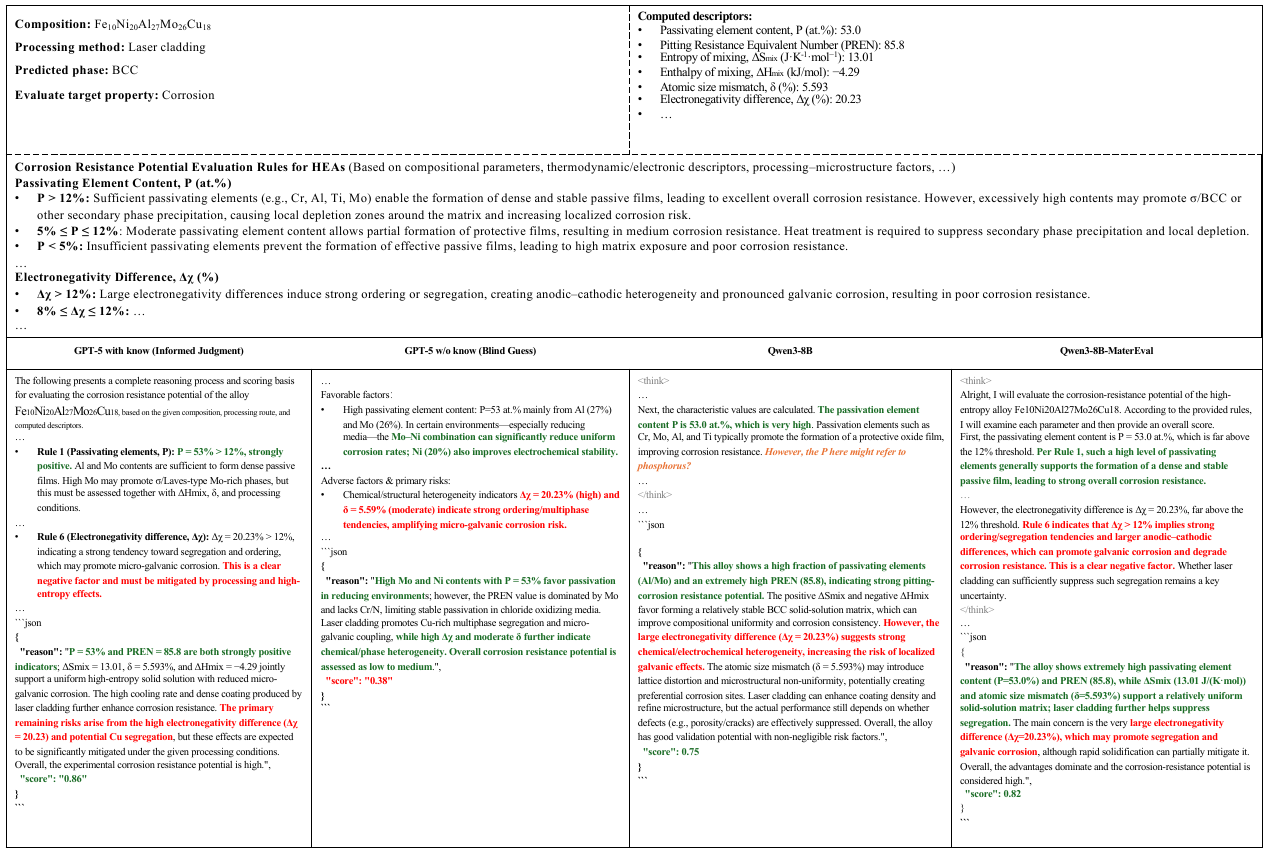}
    \caption{}
    \label{fig:corrosion_case_study}
\end{figure}

Fig.~\ref{fig:corrosion_case_study} presents a second case study focusing on corrosion-resistance evaluation, using the HEA Fe$_{10}$Ni$_{20}$Al$_{27}$Mo$_{26}$Cu$_{18}$ fabricated via laser cladding, with a predicted BCC phase. 
Compared with the phase-evaluation case, corrosion assessment involves a different set of coupled criteria, including passivating element content, pitting resistance indicators, and chemical heterogeneity, providing a complementary test of model judgment behavior.

When explicit knowledge is available, GPT-5 (\emph{with know}) performs a balanced and rule-consistent evaluation. 
It correctly identifies the extremely high passivating element content ($P=53\%$) and PREN value (85.8) as strong positive indicators for corrosion resistance, while simultaneously highlighting major risk factors such as the large electronegativity difference ($\Delta\chi=20.23\%$) and potential Cu segregation. 
By explicitly weighing these competing effects and accounting for mitigation from laser cladding, the model assigns a high but non-saturated score (0.86), reflecting an informed judgment rather than an optimistic extrapolation.

In contrast, GPT-5 operating without knowledge support (\emph{w/o know}) exhibits a markedly different behavior. 
Although it recognizes similar surface-level indicators, the absence of explicit rule grounding leads to inconsistent prioritization of evidence, resulting in a substantially lower score (0.38). 
This divergence illustrates how, even for strong models, blind guessing can cause the final judgment to drift when multiple competing factors must be reconciled.

The limitations of unaligned small models are more pronounced in this case. 
The base Qwen3-8B model not only assigns a relatively high score (0.75), but also makes a fundamental domain error by misinterpreting the passivating element content $P$ as referring to elemental phosphorus rather than a composite descriptor defined by corrosion rules. 
Such a mistake directly undermines the validity of the subsequent reasoning, highlighting that, without alignment, small LLMs may lack even basic semantic grounding of domain-specific descriptors in vertical scientific tasks.

After preference-aligned training, Qwen3-8B-MaterEval no longer exhibits this failure mode. 
The aligned model correctly interprets $P$ as the aggregated fraction of passivating elements, explicitly weighs its strong positive contribution against the large electronegativity difference, and produces a structured, evidence-grounded judgment with a high but moderated score (0.82). 
Its reasoning trajectory closely mirrors that of the informed teacher, demonstrating stable integration of multiple corrosion-related criteria under a fully rule-free inference setting.

Taken together with the phase-evaluation case in Fig.~\ref{fig:phase_case_study}, these two examples illustrate complementary aspects of material judgment. 
The first case highlights how blind guessing can lead to overly optimistic scores when destabilizing evidence is underweighted, while the second exposes how unaligned small models may fail at a more fundamental level by misinterpreting core domain descriptors. 
Across both cases, preference-aligned models consistently exhibit improved evidence integration, semantic grounding, and judgment stability, aligning their behavior with informed expert evaluation rather than heuristic-driven guesswork.

Together, these two case studies illustrate how knowledge-augmented preference alignment reshapes model behavior at the level of concrete material evaluations. Rather than promoting more elaborate explanations, the alignment process guides models to consistently integrate relevant evidence and to avoid intuitive but weakly grounded heuristics when forming material judgment decisions.

\section{Open Access and Licensing}
\label{sec:license}

The source code developed in this study is released under the Apache License, Version 2.0. 
This license permits unrestricted use, modification, and distribution of the code, provided that the terms of the license are satisfied.

The datasets constructed and used in this work are released under the Creative Commons 
Attribution–NonCommercial 4.0 International (CC BY-NC 4.0) license. The data may be used, 
shared, and adapted for non-commercial purposes with appropriate attribution, in accordance with the license terms.

All trained model weights produced in this study are fully open-sourced and publicly released under the same non-commercial usage constraints as the dataset. The models may be used for research and academic purposes and redistributed under the conditions specified by the corresponding licenses.

The code, datasets, and model weights are openly accessible via the repositories linked in the \emph{Data availability} and \emph{Code availability} sections.

\end{appendices}

%===========================================================================================%%
% If you are submitting to one of the Nature Portfolio journals, using the eJP submission   %%
% system, please include the references within the manuscript file itself. You may do this  %%
% by copying the reference list from your .bbl file, paste it into the main manuscript .tex %%
% file, and delete the associated \verb+\bibliography+ commands.                            %%
%===========================================================================================%%

\bibliography{sn-bibliography}% common bib file

\section*{Acknowledgements}

This work was sponsored by the Advanced Materials-National Science and Technology Major Project (2025ZD0620100), National Key Research and Development Program of China (No.2023YFB4606200), Key Program of Science and Technology of Yunnan Province (No.202302AB080020). 

\section*{Author contributions}
\textbf{Yeyong Yu}: Writing - Original draft, Data curation, Model training and optimization, Implementation, Methodology, Investigation, Formal analysis, Visualization. 
\textbf{Wenya Hu}:  Experimentation, Data analysis, Statistical evaluation, Model performance assessment, Validation. 
\textbf{Xing Wu}: Investigation, Writing - Review \& Editing. 
\textbf{Quan Qian}: Conceptualization, Methodology, Funding acquisition, Project administration, Supervision, Writing - Review \& Editing.

\section*{Competing interests}
The authors declare that they have no conflicts of interest/competing interests.
\end{document}